# Using deep convolutional neural networks to classify poisonous and edible mushrooms found in China

**Baiming Zhang**
Northeast Yucai Foreign Language School
Shenyang, Liaoning, China
2559467001@qq.com

**Ying Zhao** *
Northwestern University
Evanston, IL, USA
yingzhao2018@u.northwestern.edu

**Zhixiang Li**
Department of Biomedical Engineering
Shenyang Pharmaceutical University
Shenyang, Liaoning, China
106040205@syphu.edu.cn

## Abstract

Because of their abundance of amino acids, polysaccharides, and many other nutrients that benefit human beings, mushrooms are deservedly popular as dietary cuisine both worldwide and in China. However, if people eat poisonous fungi by mistake, they may suffer from nausea, vomiting, mental disorder, acute anemia, or even death. Each year in China, there are around 8000 people became sick, and 70 died as a result of eating toxic mushrooms by mistake. It is counted that there are thousands of kinds of mushrooms among which only around 900 types are edible, thus without specialized knowledge, the probability of eating toxic mushrooms by mistake is very high. Most people deem that the only characteristic of poisonous mushrooms is a bright colour, however, some kinds of them do not correspond to this trait. In order to prevent people from eating these poisonous mushrooms, we propose to use deep learning methods to indicate whether a mushroom is toxic through analyzing hundreds of edible and toxic mushrooms smartphone pictures. We crowdsource a mushroom image dataset that contains 250 images of poisonous mushrooms and 200 images of edible mushrooms. The Convolutional Neural Network (CNN) is a specialized type of artificial neural networks that use a mathematical operation called convolution in place of general matrix multiplication in at least one of their layers, which can generate a relatively precise result by analyzing a huge amount of images, and thus is very suitable for our research. The experimental results demonstrate that the proposed model has high credibility and can provide a decision-making basis for the selection of edible fungi, so as to reduce the morbidity and mortality caused by eating poisonous mushrooms. We also open source our hand collected mushroom image dataset so that peer researchers can also deploy their own model to advance poisonous mushroom identification.

## 1 Introduction and Background

Mushroom poisonings happen when forager misidentify a poisonous species edible. In most cases, mushroom poisonings have benign symptoms of generalized gastrointestinal upset. However, the poisonings may also lead to severe symptoms including liver failure, kidney injury and neurological sequelae or even risk of death. [1] According to the foodborne disease outbreak surveillance system from China Center of Disease Control (CCDC), within the decade of 2010 to 2022, a total of 10,036 mushroom poisoning outbreaks were detected in China. These outbreaks resulted in 38,676 illness and 788 deaths. Mushroom poisoning outbreaks also exhibited seasonality, with the peak season occurring between May and October and accounting the highest prevalence of illness and death. [2] CCDC called on the targeted health reduction on the consumption of mushrooms to the general population, as well as advised not to hunt and consume wild mushrooms. Therefore, we planned to develop a deep learning-based classifier to identify the poisonous mushrooms from mushroom images. Such an image classifier is expected to reduce the potential risk of mushroom poisoning.

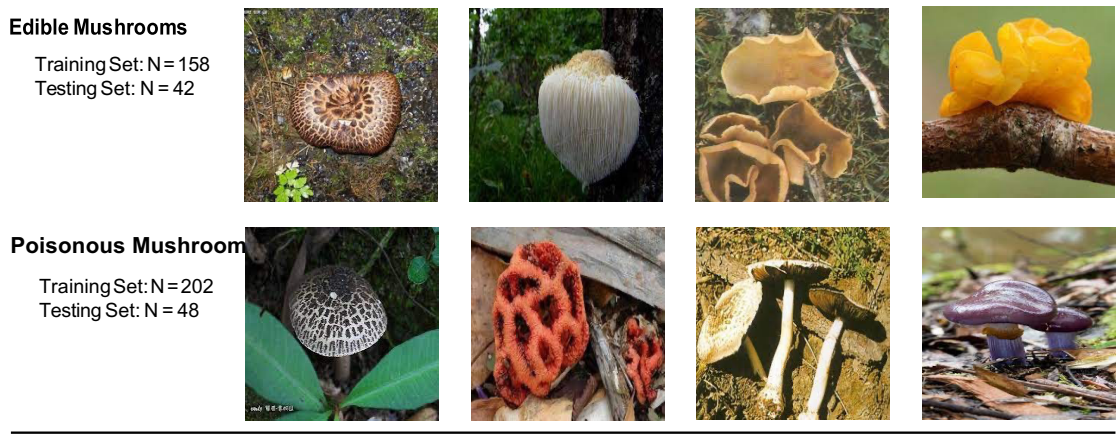

Figure 1: Summary statistics and examples of the fungi dataset

Table 1: Performance of 4 deep convolutional neural networks on the classification of poisonous mushrooms

| Model | Accuracy | AUC | Precision | Recall | F1 |
|---|---|---|---|---|---|
| *AlexNet* | 0.700 | 0.748 | 0.705 | 0.750 | 0.727 |
| *VGG16* | 0.689 | 0.735 | 0.689 | 0.833 | 0.755 |
| *DenseNet121* | 0.744 | 0.764 | 0.756 | 0.708 | 0.731 |
| *ResNet50* | 0.756 | 0.809 | 0.782 | 0.750 | 0.766 |

Thanks to the fast development of computational technology, deep learning and machine learning has been widely applied to different aspects. Image classification tasks also benefits a lot from the emerge of deep convolutional neural networks such as automatic disease identification [3], development of autonomous car [4], motion detection [5], facial recognition [6] and etc. Particularly in mushroom classification, Ketwongsa et al. [7] applied AlexNet to classify the poisonous and edible mushrooms from the five most commonly species found in Thailand. Gupta et al. [8] also employed convolutional neural networks to classify whether a given mushroom in the Agaricus and Lepiota Family Mushroom is edible or poisonous. Zahan et al. [9] also used convolutional neural networks to classify if a mushroom is edible, inedible and poisonous mushroom classification. Unlike prior studies, our study is novel in that 1) we only used the mushroom species that found in China 2) our mushroom dataset was directly collected from online sources instead of the high-quality and find-resolution images used in literature. Therefore, our proposed models are expected to be more generalizable and practical if used in China.

Our study has the following two contributions:

1. We crowd-sourced and released a public dataset that contains the images of 250 poisonous mushrooms and 200 edible images.

2. We developed deep convolutional neural network-based classifiers to identify the poisonous species from mushroom images

## 2 Methods

### 2.1 Dataset

We collected an image dataset by crowd-sourcing the fungi images found in China from the internet. The edible mushroom images are selected with reference to the Chinese Edible Fungi Checklist [10] released by the Ministry of Health of the People's Republic of China . The poisonous mushroom images are selected based on a revised checklist introduced by Bau et al. [11]. We searched and downloaded 200 images of Chinese edible mushrooms and 250 images of Chinese poisonous images based on their scientific names. The example images are shown in Figure 1. No mushrooms of duplicated species can be found in the dataset. We split the entire dataset into training (N=360) and hold-out testing set (N=90). We open-source this dataset so that those peer researchers who are also interested in mushroom classification can benefit from our resources. For public access of this dataset, please go to: https://drive.google.com/drive/folders/13NFDI5UhcLHPSL2WMcOrFs6QhhmrDjxQ?usp=sharing

### 2.2 Pre-processing

The images should be pre-processed before feeding into the deep learning networks. We sequentially applied the following standard pre-processing steps. For all images, we first resized each image to the size of 256 by 256. We then randomly cropped all images in the training set to the size of 224 by 224. For the holdout testing set, we cropped them to the same size but using a central cropping mechanism. We also randomly horizontally flip the images in the training set for data augmentation. Finally, all images are normalized using the (mean, standard deviation) pairs of [(0.485, 0.229), (0.456, 0.224), (0.406, 0.225)] for each of the RGB channels.

### 2.3 Deep Learning Models

Deep convolutional neural networks are widely used in the image classification tasks. In our study, given our limited sample size of our datasets we applied the deep learning models pre-trained on ImageNet. We simply replaced the last fully connected layers of those pre-trained models to a linear layer that have an output dimension of 2 to produce the probability for each of the binary outcomes, edible or poisonous. We loaded the pre-trained weights and inherited their model architecture of the following models: AlexNet [12], VGG16[13], DenseNet121[14] and ResNet50[15]. For the technical details. We used stochastic gradient descent (SGD) optimizer with learning rate 0.001 and a momentum of 0.9. As our problem is a binary classification, we employed binary cross-entropy as our loss function. The batch size was set to be 4. We trained each classifier for 500 epochs with early termination conditioned on the loss of validation set. In other words, if the model performance on the validation set did not improve in 5 consecutive epochs, we stopped the training process and evaluated our models on the testing set. We compared the performance of these legendary pre-trained image models.

### 2.4 Evaluation

The mushroom classification can be cast as a binary outcome classification. Given that the dataset is slightly imbalanced, solely using the accuracy to evaluate model performance is not reliable enough. Besides, accuracy, we also report area under the receiver operating characteristic (AUC), precision, recall and F1-score. To select best-performing hyper-parameters, we further left out a validation set (N=40) from the training set. The hyper-parameters achieving the best performance evaluated by accuracy will be used in the held-out testing set. We repeated each experiment 5 times and report the mean value for each of the evaluation metrics.

### 2.5 Implementation

We build the entire pipeline (as shown in Figure 2) using Python 3.8. PyTorch and TorchVision is used to design deep neural networks and load pre-trained models. Pandas and Numpy is applied to pre-process data. Google Colab with a GPU runtime is used to reduce training time. Scikit-learn package is used to evaluate the model performance. Each epoch took approximately 5 seconds to run.

## 3 Results

The results are shown in Table 1. The ResNet50 model achieved the best results of accuracy 0.756, AUC 0.809, precision 0.782, recall 0.750 and F1 0.766, followed by DenseNet121 (accuracy 0.744, AUC 0.764, precision 0.756, recall 0.750 and F1 0.731). The VGG16 and AlexNet model yielded sub-optimal results when compared to the above two models. Our models have the feasibility in practical uses in that they produce high recall values, which means our models are sensitive to the poisonous mushrooms. Those mushrooms have a high probability of poisoning might be red-flagged by our models, which will remind the potential users to avoid eating them.

## 4 Limitation and Conclusion

Our work also has following limitations. First, our sample size is relatively small. A larger dataset could potentially improve our model performance. We plan to try using the automatic web scraping tools to download more mushroom images to enlarge the dataset. In addition, we don't apply some other trending image classification model such as deep InceptionV3 [16] or SwinTransformer [17], which have shown great successes in some downstream image classification tasks. We plan to integrate these algorithms to our experiments should there be any computational resources available.

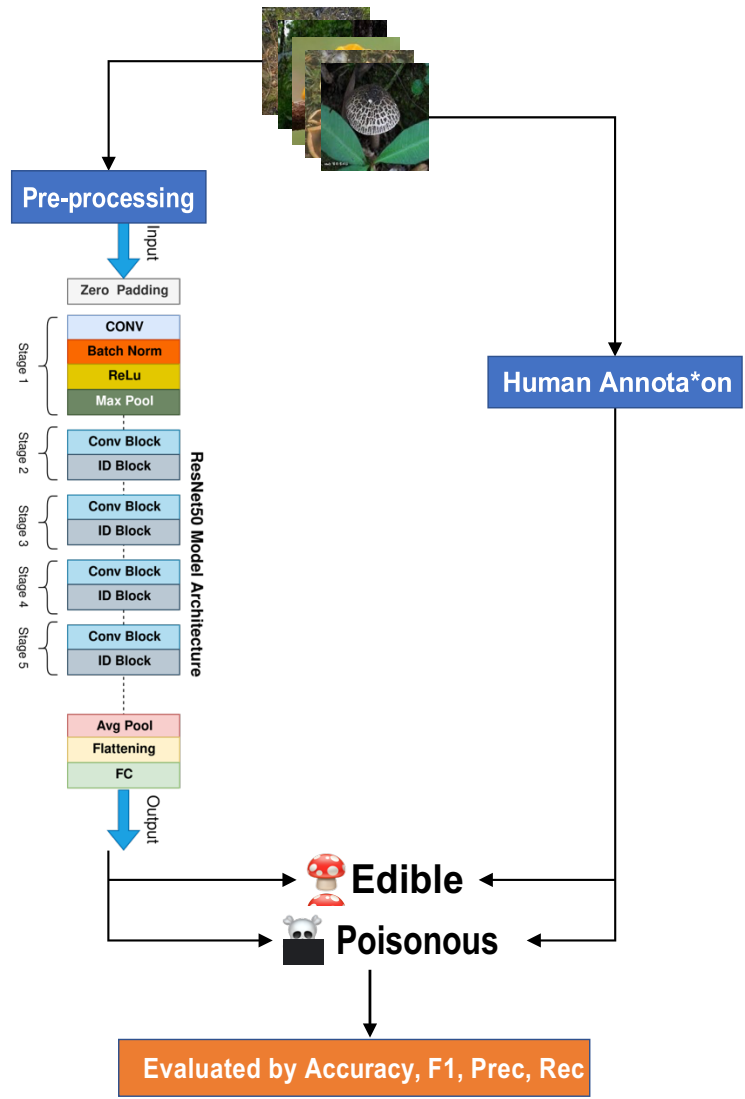

Figure 2: Pipeline of the proposed study

Moreover, our proposed models cannot provide more information besides the toxicity of a mushroom. We plan to integrate more functions, such as mushroom identification at the next step. In the future study, we will keep our models and datasets updated and embed our models to a front-end application so that end-users can benefit from these new technologies to reduce the risk of mushroom poisoning.

## 5 Author Contribution Statement

All authors discuss and define the research problem. B.Z. extracts and pre-process all the research data. B.Z. and Z.L. implement all deep neural networks, analyze and visualize the results. B.Z. and Y.Z. write the manuscript advised by Z.L., who also organizes the manuscript to a better academic standard.